\def\year{2017}\relax
\documentclass[letterpaper]{article} 
\usepackage{ijcai18}  
\usepackage{times}  
\usepackage{helvet}  
\usepackage{courier}  
\usepackage{url}  
\usepackage{graphicx}  
\usepackage[dvipsnames]{xcolor}
\usepackage{amsopn}
\usepackage{amsmath}
\usepackage{amsfonts}
\usepackage{amssymb}
\usepackage{algorithm}
\usepackage[noend]{algpseudocode}
\usepackage{subcaption}
\usepackage{booktabs}
\usepackage{multirow}
\usepackage{amsthm}
\usepackage{textcomp}
\usepackage{bm}
 \usepackage[font={small}]{caption}

\DeclareMathOperator*{\argmin}{arg\,min}

\theoremstyle{definition}
\newtheorem{defn}{Definition} 
 
\makeatletter
\def\BState{\State\hskip-\ALG@thistlm}
\makeatother
\begin{document}
%
\newcommand{\sscomment}[1]{\textcolor{red}{[Sid] #1}}

\title{Model-Free Model Reconciliation}
%
%
%
%
%
%
%
%
\author{
Sarath Sreedharan, Alberto Olmo, Aditya Prasad Mishra \and Subbarao Kambhampati \\
School of Computing, Informatics, and Decision Systems Engineering\\
Arizona State University, Tempe, AZ 85281 USA\\
{ \{ ssreedh3, aolmoher, amishr28, rao  \} @ asu.edu}}
\maketitle
\begin{abstract}
Designing agents capable of explaining complex sequential decisions remain a significant open problem in automated decision-making. 
Recently, there has been a lot of interest in developing approaches for generating such explanations for various decision-making paradigms.
One such approach has been the idea of {\em explanation as model-reconciliation}.
The framework hypothesizes that one of the common reasons for the user's confusion could be the mismatch between the user's model of the task and the one used by the system to generate the decisions.
While this is a general framework, most works that have been explicitly built on this explanatory philosophy have focused on settings where the model of user's knowledge is available in a declarative form.
Our goal in this paper is to adapt the model reconciliation approach to the cases where such user models are no longer explicitly provided.
We present a simple and easy to learn labeling model that can help an explainer decide what information could help achieve model reconciliation between the user and the agent.
\end{abstract}

\section{Introduction}
A major barrier to integrating AI systems into our daily lives has been their inability to interact and work with us in an intuitive and explicable manner. Orchestrating such interactions would require the agents to have the ability to help users in the loop better understand the rationale behind their various actions. Thankfully there has been a lot of interest within the AI research community to develop systems capable of holding explanatory dialogue with users and thus help them understand the decisions under question \cite{miller,danmaga}. When a user is stumped by an agent's decision and asks for explanation, two common reasons as to why the user may have trouble understanding the agent's actions could be due to either (1) the user's lack of understanding (or even misunderstanding) of the task or (2) because of the user's inability to understand the consequences of the actions due to their limited inferential capacity. While many earlier works in explanation have generally focused on the latter (c.f \cite{khan2009minimal,hayes2017improving,seegebarth,gen-pol}), there is a growing consensus on the importance of explanatory mechanisms that can help bridge the knowledge asymmetry between the system and the user. 

In particular, this problem has been studied within the context of classical planning and has been referred to as \textbf{explanation as model-reconciliation} \cite{explain}. Most works in this direction have generally looked at cases where the user's model of the task (i.e their belief about the initial state, the transition dynamics, and the goal) is known beforehand (in a representation scheme comparable to the one used by the agent) and do not match the agent's model. This mismatch means that the user would not be able to correctly evaluate the validity or the optimality of a given plan. Thus their explanations consist of information about the agent's model that the user could incorporate into their own model to correctly evaluate the plan in question.

\begin{figure}[htp]
\centering
\includegraphics[scale=0.37]{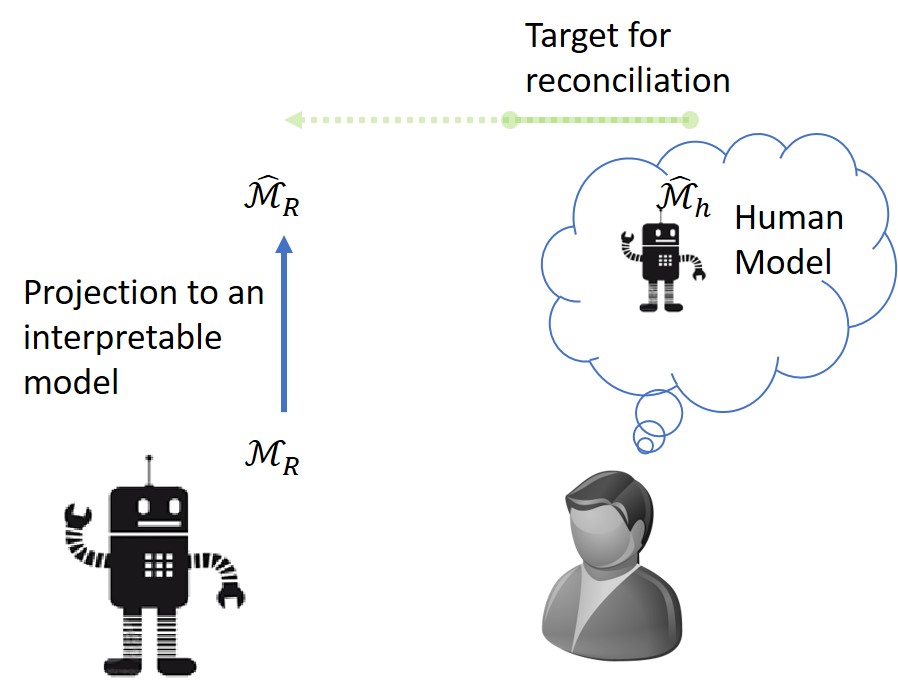}
\caption{A general overview of the explanation as model reconciliation.}
\label{fig:1}
\vspace{-5pt}
\end{figure}
Unfortunately, it is not always possible to have access to such models. In the most general case, we are dealing with user's model of the agent and hence the user may not be capable of presenting traces or decisions that could be generated from this model. 
Even if the system tries to learn such a representation based on interactions with the user, there is no guarantee that the specific representational assumptions of the learned model and the vocabulary used would be satisfied by the user's mental model.

The definition of explanation as model reconciliation may leave one with the idea that there is no way around it. How could one ever truly perform effective reconciliation when there exists no user model guiding us to choose the parts of the model, which when revealed to the user will help them correctly evaluate the current decision? Are we left with revealing the entire agent model to the user as the only option? In this paper, we propose a simple and intuitive way we could still generate minimal explanations in the absence of declarative models. We argue that we could drive such explanations by using simple and easy to learn models that can predict how human expectations could be affected by possible explanations (derived completely from information about the agent model) and in fact show how this method could be viewed as a variation of previous approaches that have been put forth to identify explicable behavior.

We will ground this discussion within the context of MDPs and start by summarizing how model reconciliation could be achieved in such settings when the user model may be known (Section \ref{sec:formulation}). The rest of the paper will investigate how these ideas could be used when the human mental model of the task is unavailable,and will formulate a learning problem that allows us to learn simple models that could be used to identify minimal explanations (Section \ref{sec:missing_models}). We will evaluate our method on a set of standard MDP benchmarks and perform user studies to validate its viability.
\vspace{-5pt}
\section{Background}
Figure \ref{fig:1}, presents a general schematic representation for explanation as model reconciliation. The automated agent (henceforth referred to as robot) starts with a model $\mathcal{M}_R$ for coming up with a decision $\pi$ (where depending on the context, $\pi$ may be a single action, plan, policy or a label). The model is then projected into a space of interpretable models $\hat{\mathcal{M}}_R$ such that some desirable properties of the decision (say the optimality) may be conserved. In this setting, $\hat{\mathcal{M}}_H$ corresponds to the human's preconceived notions about the robot model captured in the same interpretable space. The explainer's job then becomes providing information about the model $\hat{\mathcal{M}}_R$, such that the updated human model can correctly evaluate the validity of the robot decisions. Note that in this case, the robot could have chosen to provide the entire model, but for most realistic tasks, such models could be quite large, so dumping the entire model could be both unnecessary and impractical. It's also well known that people generally prefer explanation that are in line with their beliefs \cite{miller}. Thus the users would be happier with explanations that asks them to update a subset of beliefs as opposed to a complete update.

In \cite{explain}, $\mathcal{M}_R$ was a classical planning model hence inherently interpretable and thus the reconciliation is performed with the original domain model. This idea could be applied beyond just planning models, for example, one could understand the explanation methodology used by LIME \cite{ribeiro2016should} as being a special case of model reconciliation. In their case, they assume the human model is empty and $\hat{\mathcal{M}}_R$ is automatically generated for each decision using a set of predefined features.

In this work, we will be looking at the agents that use discounted infinite horizon Markov Decision Processes (or MDPs) \cite{russell2003artificial} as the decision making framework.
Each MDP $\mathcal{M}$ is defined by a tuple $\langle S, A, T, R, \gamma, \mu\rangle$, where the $S$ provides the set of possible atomic states, $A$ defines the set of actions, $T$ is the transition function, $R$ the reward, $\gamma$ the discounting factor (where $0\leq \gamma < 1$) and $\mu$ corresponds to the distribution of possible initial states.
$T: S\times A\times S \rightarrow [0,1]$ provides the probability that for given state $s \in S$, the execution of an action $a$ would induce a transition to a new state $s'$, and $R: S\times A\times S \rightarrow \mathbb{R}$ defines the reward corresponding to this transition. 
The solution concept in MDP takes the form of a policy $\pi$ that maps each state to a potential action. 
A policy is said to be optimal for $\mathcal{M}$ (denoted as $\pi^*_{\mathcal{M}}$) if there exist no other policy that dominates the given policy in terms of the expected value of states. Executing the policy in a state results in a sequence of state action state tuples called {\em execution trajectory} or simply a {\em trajectory}, denoted as $\tau = \langle (s_1, a_1, s_2),..., (s_{n-1}, a_{n-1}, s_{n})\rangle$ and we will use $P_{\mathcal{M}}(\tau|\pi)$ to denote the probability of sampling the given trajectory $\tau$ for a policy $\pi$.

In the explanatory setting we are interested in, robot uses a model $\mathcal{M}^r = \langle S, A, T^r, R^r, \gamma^r\rangle$ of the task to come up with the policy to act on.
For now we will assume this MDP already defines an interpretable model and the human uses a model $\mathcal{M}^h = \langle S, A, T^h, R^h, \gamma^h\rangle$ to evaluate it (we will relax this assumption in later sections). Now the task ahead of us will be to reinterpret the ideas of inexplicability and the idea of model reconciliation that was defined in \cite{explain} into the context of MDP, but before we go into the detail, let us consider a simple explanatory scenario.

\vspace{-5pt}
\section{Illustrative Example}
Consider a warehouse scenario, where a robot worker is tasked with moving packages from racks and dropping them off at the dispatch chute. The robot is powered by a battery pack that can be recharged by visiting its docking station. The docking station also doubles as a quality assurance station that the robot needs to visit whenever it picks up a box labeled \#013 (which means the box is fragile). The robots operations are mostly deterministic, apart from a small probability of slipping (0.25) in some cells, that could leave the robot in the same position.

Now suppose the warehouse has just hired a new part-time employee to oversee the operations. The employee is just getting used to this new setting and is puzzled by the robot's decision to once in a while take a detour from the drop-off activity and visit a specific position of the factory floor (which is, in fact, the docking location). If we wished the robot to be explainable, then it would need to be capable of helping the employee better understand the characteristics of its operations (i.e achieve some form of model reconciliation). Given the fact that the robot may not have an exact model of the user, one way to achieve this could be by providing robot's entire model to the user. Unfortunately, this could easily overwhelm the user. 

Another possibility could be to allow the user to specify which robot actions appear inexplicable, and focus on providing facts relevant to those actions. This explanation may still prove to be quite verbose and may in fact not help resolve their confusion. 
For example, imagine a case where the robot is visiting the station to recharge its batteries and the human says that the visit action is inexplicable.
Now even if the robot mentions that visiting the station recharges it, the employ may still be confused if they are under the incorrect assumption that the robot is operating on full battery.
Similarly, if the human had expected the robot to go to the docking station due to some confusion regarding the box codes, the human may mark the robot decision to not go to the dropoff as being inexplicable and the explanations that could resolve the confusion may have little to do with the specific action marked as inexplicable. 

\begin{figure*}[htp]
\centering
\includegraphics[scale=0.4]{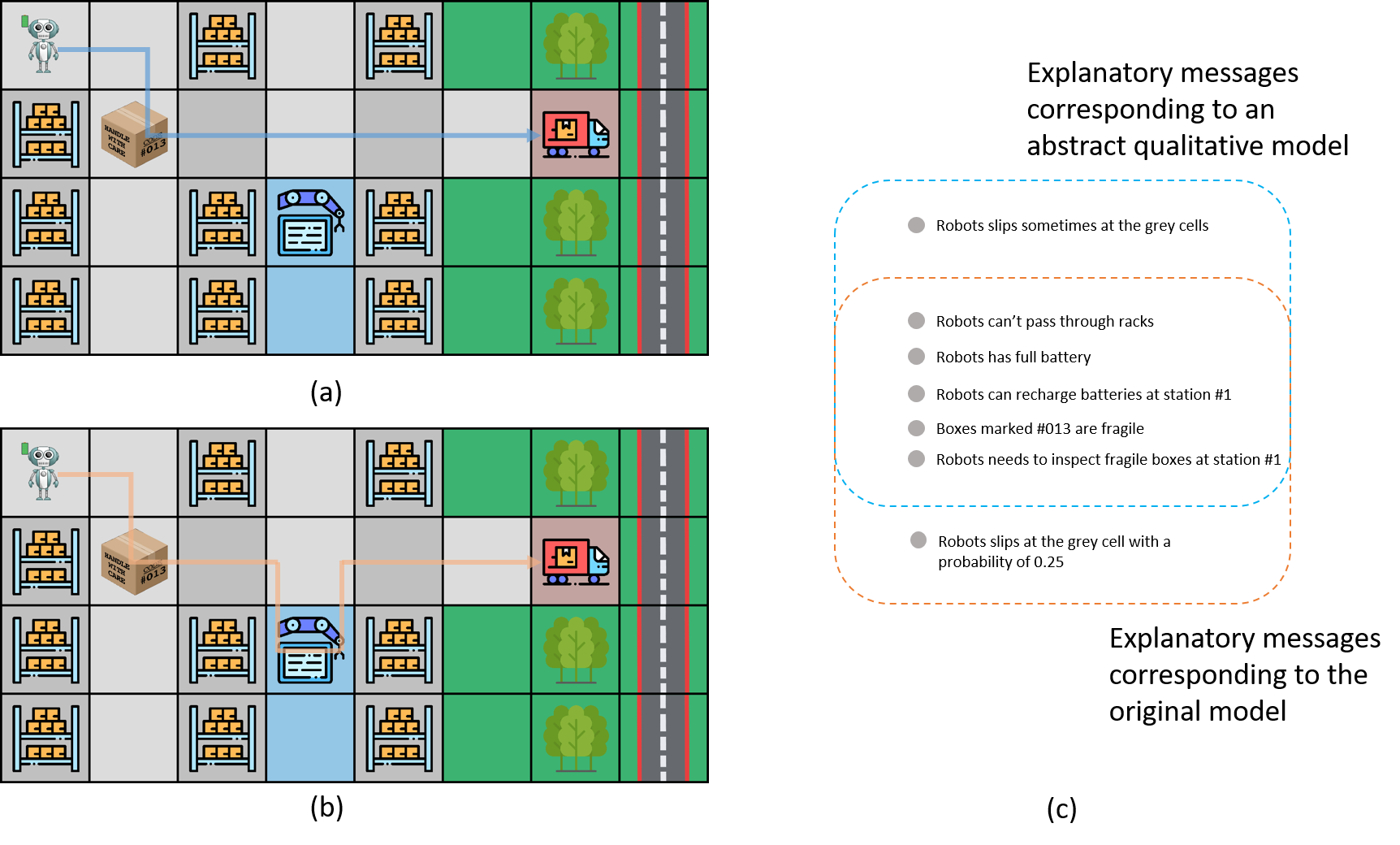}
\caption{Subfigure (a) shows a visualization of a trajectory expected by the user described in the illustrative example, and (b) shows the visualization of a trajectory the user may observe. Subfigure (c), shows the various explanatory messages that could be used in this scenario, note that the messages span information from multiple abstractions of the given task}
\label{fig:2}
\vspace{-12pt}
\end{figure*}
\section{Explanation as Model Reconciliation For MDPs}
\label{sec:formulation}
In this paper, we will assume that the human and robot models (captured as MDPs) differs solely on the transition probabilities, the reward function and the discounting factor. This means that we can characterize both models by the tuple $\theta(\mathcal{M}) =\langle \theta_{T}, \theta_{R}, \theta_{\gamma}, \theta_{\mu}\rangle$, where the $\theta_{T}$ provides the set of parameters that defines the transition probabilities $P(.|s,a)$, while $\theta_{R}$ the parameters corresponding to the reward function,  $\theta_{\gamma}$ the parameters corresponding to the discount factor and $\theta_{\mu}$ the parameters for the initial state distribution. For simple MDP models with atomic states, $\theta_{T}$ contains parameters of a categorical distribution for each transition ($\theta_{\mu}$ will contain similar parameters), $\theta_{R}$ contains the reward associated with each transition (an $\langle s,a,s'\rangle$ tuple) and $\theta_{\gamma}$ just contains the value of the discount factor.
For simplicity, we will denote each of the unique parameters in the larger set using indexes. For example, $\theta_{T}^{s,a}$, will correspond to the parameters for the distribution $P(.|s,a)$.

Mathematically, a model reconciliation operation will be captured as a function $\mathcal{E}_{\langle \mathcal{M}^h, \mathcal{M}^r\rangle}: 2^{\Theta} \rightarrow \mathbb{M}$ that takes in a set of model parameters and generates a new version of the model $\mathcal{M}^h$ where the set of specified parameters will be set to values from $\mathcal{M}^r$. For example,  $\hat{\mathcal{M}} = \mathcal{E}_{\langle \mathcal{M}^h, \mathcal{M}^r\rangle}(\theta_{T}^{s_1,a})$ will be a new model such that $\theta(\hat{\mathcal{M}})$ will be identical to $\theta(\mathcal{M}^{h})$, except that $\theta_{T_{\hat{\mathcal{M}}}}^{s1,a}$, will be equal to $\theta_{T^r}^{s1,a}$.

Practically, the model reconciliation operation corresponds to the robot informing the human about some part of its model. We can assume that this communication incurs some cost and we can define a cost function $\mathcal{C}: 2^{\Theta} \rightarrow \mathbb{R}$ that maps a given set of a threshold to a cost.

Now the question we need to ask is whether the agent is trying to explain its policy or if it is trying to explain some behavior (i.e an execution trace). Most of the earlier works that looks at model reconciliation explanation (c.f \cite{explain,sreedharan2018handling,sreedharan2018hierarchical}) have looked at sequential plans, has generally ignored this differentiation and treated the problem of explaining plans to be same as that of explaining behavior. 
In general, a given plan or policy compactly represents a set of possible behaviors and the choice of explaining behavior {\em vs} explaining the plans/policies could affect the content of the explanation being given. 
For example, when explaining policies there is the additional challenge of presenting the entire policy to the user and the explainer may need to justify action choices for extremely unlikely states or contingencies. On the other hand, when explaining a given set of behaviors the explainer needs to only justify their action choices for cases they actually witnessed.
For example, when explaining traces from the warehouse scenario, given the small probability of slipping the robot may never have to mention what to do when it slips, but on the otherhand if we are dealing with full policies, the agent may need to talk about the states where the robot is in the slipped positions and they need to get up from that position and move on.

Explaining policies or plans becomes more relevant when we consider explanatory dialogues where the agent and the user are trying to jointly come to agreement on what policy/plans to follow (eg: decision support systems), while the latter may be more useful when the user is observing some agent operating in an environment.

With respect to policies, we assume that the user is presented with the entire policy and a given policy and is said to be explicable, if the policy is optimal for the given human model. Therefore the goal of the explainer becomes that of ensuring the optimality of the given policy

\begin{defn}
{\em A set of parameters $\theta_{\mathcal{E}}$ corresponds to a \textbf{complete policy explanation} for the given robot policy $\pi_{\mathcal{M}^r}^{*}$, if the policy is also optimal for $\mathcal{E}_{\langle \mathcal{M}^h, \mathcal{M}^r\rangle}(\theta_{E})$ and is said to be the minimally complete policy explanation if there exists no other complete explanation $\theta_{\mathcal{E}'}$, such that, $\mathcal{C}(\theta_{\mathcal{E}'}) < \mathcal{C}(\theta_{\mathcal{E}})$}
\end{defn}
Finding a complete policy explanation is relatively straight forward (the set of all parameters automatically meets this requirement). The more challenging cases becomes that of finding the minimal or the cheaper explanations. 

For explaining behavior, we will look at the simplest case, namely the agent needs to explain a set of behavior that the user has just observed. We will assume that the observer has full observability of the state and is seeing the robot behavior for the first time.
In such a setting, a given trace $\tau$ would appear explicable to the user if it could be sampled from the given MDP policy or more generally, i.e $P_{\mathcal{M}^H}(\tau|\pi) > \delta$, where $\delta$ is some small threshold \footnote{We use $\delta$ instead of zero to allow for the possibility that people are often surprised by unlikely events of non-zero probability}

\begin{defn}
{\em A set of parameter $\theta_{\mathcal{E}}$ corresponds to a \textbf{complete behavior explanation} for a set of traces $\mathbb{T} = \{\tau_1, ... \tau_n\}$, if $\forall \tau \in \mathbb{T},~ \exists \pi$ such that $ P_{\mathcal{E}_{\langle \mathcal{M}^h, \mathcal{M}^r\rangle}(\theta_{E})}(\tau|\pi) > \delta$ and $\pi$ is an optimal policy for the model $P_{\mathcal{E}_{\langle \mathcal{M}^h, \mathcal{M}^r\rangle}(\theta_{E})}$. The explanation is said to be the minimally complete behavior explanation if there exists no other complete explanation $\theta_{\mathcal{E}'}$, such that, $\mathcal{C}(\theta_{\mathcal{E}'}) < \mathcal{C}(\theta_{\mathcal{E}})$} 
\end{defn}
Note that given the above definition, it may not always be possible to find a complete explanation, as the trace may genuinely contain low probability transitions.

While model reconciliation could be an important component of either policy or behavior explanation, the applicability of the model reconciliation explanations on its own for policies are limited by the fact that in all but states with the smallest state space the user would have just going over the entire policy. Thus in these setting would need to also utilize policy approximation methods, then allow users the ability to drill down on policy details as required. Since one of our goals is to focus on developing approaches that allow us to generate model reconciliation explanations without explicitly defined user models, the rest of the paper will mostly focus on behavior explanation. In section \ref{conclusions}, we will have a brief discussion on how these methods could potentially be extended to policy explanation scenarios.

\section{Explaining Without Access to Human Mental Models}
\label{sec:missing_models}
Now we will look at how we could identify cheaper complete behavior explanations when the human model is unknown. We will go one step further from identifying not only the parameters that need to be explained, but also capturing the right modality/abstractions to present the information about the parameters. That is we will no longer assume that the human is using a full MDP model to come up with their decisions.
Instead, we will assume that the robot has access to a set of explanatory messages $\mathbb{M} = \{m_1, m_2, ..., m_n\}$ that can be presented to the user. Where the messages corresponds to a set of parameter values (the parameters corresponding to a set of messages $\{m_1, ..m_k\}$ is denoted as $\mathcal{E}(\{m_1, ..m_k\})$) of the model as captured in some abstraction of this model and has a corresponding cost associated with it $\mathcal{C}$.
The abstractions to consider may depend on the specific scenario and the previous information about the intended users (laypeople vs. experts).
Some simple possibilities may be to consider qualitative models (say non-deterministic ones instead of stochastic) and considering state abstractions the given task. 
Note that, technically $\mathcal{E}(\mathbb{M})$, need not span the set of all possible model parameters, but could rather be limited to a subset of parameters has identified to be relevant to the given problem. One possible way may be to consider variations of explanation techniques like MSE \cite{khan2009minimal} to identify set of possible factors that affect the optimality of each action. 
In Figure \ref{fig:2}, the subfigure (c) shows a set of possible explanatory messages for the warehouse domain, that consists of each parameter mapped to some english statement. For models captured using factored representations that use relational or propositional variables, such statements could be easily generated using templates (c.f \cite{hayes2017improving}). 

Given this setting, we will now make two simplifying assumptions, namely,(1) the order in which the explanatory messages are presented does not matter, (2) We have access to a set of observers with similar models, and they share this model with the target user.

Now our goal is to learn a predictive model that is able to predict whether a given user would find a given $\langle s,a,s\rangle$ tuple explicable and how the user's perception changes with the given explanatory messages. 

For example, at the beginning of an episode the user may be presented with the following explanatory messages, $\hat{\mathbb{M}} = \{m_1 = \textrm{``Robot slips with probability 0.25 at grey cells"}\}$, which corresponds to the fact that $P(s_i|a, s_i) = 0.25$, for all states $s_i$ where the feature grey cell is true and for all actions $a$.
Now the user will be presented with a sequence of transitions, say $\langle (1,2), \textrm{right}, (2,2)\rangle$ and asked whether the transition was explicable or not. Then the tuple $\langle\langle (1,2), \textrm{right}, (2,2)\rangle, \{m_1\}, l_1\rangle$, where $l_1$ is the label assigned by the user to the transition, becomes input to our learning method.

The exact function, we would want to learn would be
\begin{equation*}
\mathcal{L}(\langle s,a,s'\rangle,\{m_1, ...,m_k\}) = \begin{cases}
1 &\text{if }  \langle s,a,s'\rangle \sim \\ & \pi^{*}_{\mathcal{E}_{\langle\mathcal{M}^h,\mathcal{M}^r\rangle}(\{\theta_{\psi_1},..,\theta_{\psi_k}\})}(s_0)\\
0 &{\small \text{otherwise}}
\end{cases}
\end{equation*}
Note that this is a modified version of the sequential model introduced in \cite{exp-yu} for identifying whether a given plan is explicable or not. Though our methods vary in some significant aspects, namely, (1) We allow for the possibility that the explicability of the actions/traces could be affected by explanations provided by the system; (2) We no longer use labels of high level tasks as a proxy for the explicability of the trace. Instead, we just use a simple binary label on whether the transition is explicable or not; (3) We no longer consider sequence models but rather a much simpler labeling model that maps a single transition to the explicability label. We argue that in cases where the human is markovian on the same set of features as the agent, this rather simpler model suffices.

It is also important that our learning approach is more principled than the ones studied in \cite{exp-yu}, since in their case to build a balanced dataset (of explicable and inexplicable plans), they would need to uniformly sample through the entire plan space (an extremely hard endeavour with no obvious known approaches), while we stick to traces generated from the optimal policy and only need to randomly generate possible sets of explanatory messages, which is clearly a smaller set. 

Once we have learned an approximation of the above labeling function $\hat{\mathcal{L}}$, the problem of explanation generation for a trace $\tau$ becomes that of finding the subset of $\mathbb{M}$ that balances the cost of communication with the reduction in the inexplicability of the given trace, i.e
\[\argmin_{\hat{\mathbb{M}}} (\mathcal{C}^{\mathcal{M}}(\hat{\mathbb{M}})+\alpha * \Sigma_{i=0}^{n}(1- \hat{\mathcal{L}}(\langle s,a,s'\rangle,\hat{\mathbb{M}}))\]
Where $\hat{\mathbb{M}}$ is a subset of $\mathbb{M}$ and $\alpha$ is some scaling factor that balances the cost of explanation with that of number of inexplicable transitions for a given trace.
\vspace{-5pt}
\section{Evaluation}
For evaluation, we validated our approach on both simulations and on data collected from users.
For simulations, we used slightly modified versions of the Taxi domain \cite{dietterich1998maxq} (of size 6*6), the Four rooms domains \cite{sutton1999between} (of size 9*9) and the warehouse scenario (of size 9*9) described before.
For each domain, we start with an MDP instance (henceforth referred to as the robot model) and then create a space of possible user models by identifying a set of possible values for each MDP parameter. For example, in the taxi domain the parameters include position of the passengers, their destination, the step cost, discounting etc.., for the Four rooms this included the goal locations, locations with negative rewards, discounting, step cost, slip probability, etc.., and finally for the warehouse, the position of the box, the position of station \#1, the step cost, slipping probabilities and the discounting factors were selected as potential parameters that can be updated.
In this setting, we assume that there exists a single explanatory message for each possible parameter.

For each individual test, we select a random subset of three parameters and then randomly choose a value for each of these. We then treat this new MDP model as a stand-in for the user model and use it to label traces generated from the original MDP.
The traces were generated by choosing a random initial state and then following the optimal policy of the robot untill either the terminal state is reached or the trace length reaches a predefined limit.
For each trace, a random subset of the explanations was selected and presented to the human. This means updating the MDP parameters to their corresponding values in the robot model only for the parameters specified by the current subset of explanation.
Each individual transition was then labeled using this updated MDP. A transition was labeled as inexplicable if the action is not the optimal one in the human model (i.e. Q value is lower) or the next state has a probability of occurring was $\delta=0$.
\begin{figure}[htp]
\centering
\includegraphics[scale=0.2]{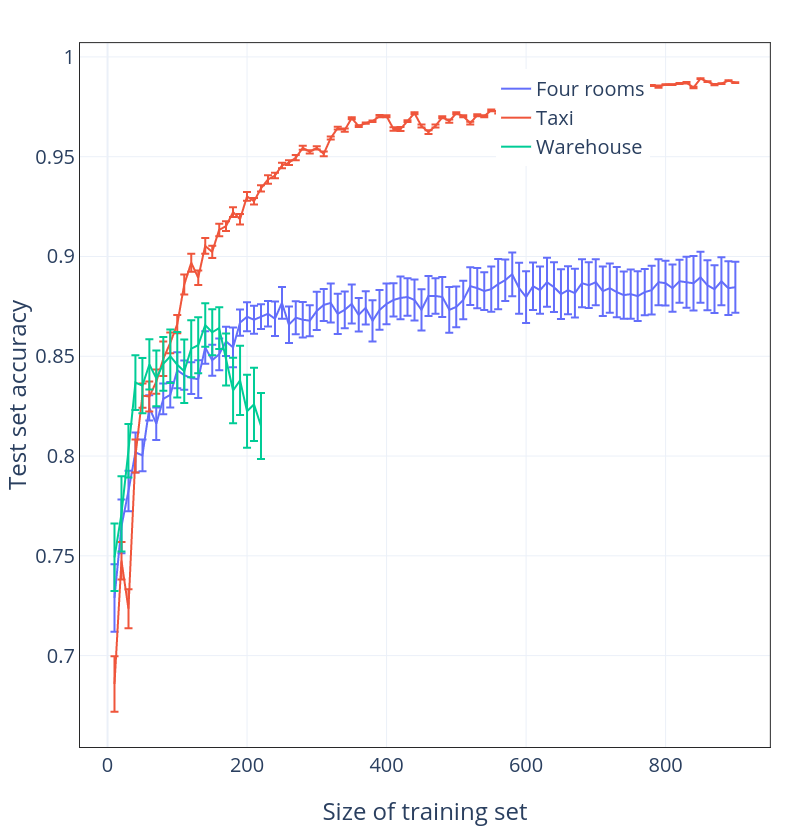}
\caption{The test accuracy for increasing sizes of training set.}
\label{fig:3}
\vspace{-10pt}
\end{figure}

We then used this set of labeled transitions to create a training set and test set for a decision tree.
The input features to the decision tree consist of current state features, (just x and y for Four rooms and the position of the the taxi and passengers for the Taxi domain and for Warehouse it included the position of the robot and the fact whether the agent picked up the box or visited station \#1), the index of the action and features capturing the current subset of explanations being considered.
In each Warehouse and Four rooms test instance, we collected 900 unique data points as training set and 100 data points as the test set. Due to the complexity of the taxi domain, we generated less data points (since for each different explanation subset we need to solve a new planning problem) and used close to 220 unique points as training data and on average 28 data points as the test set.

We then tested on 20 such instances for each domain. Figure \ref{fig:3} plots the average test accuracy for models trained with training sets of varying sizes.
As evident from the graph, a simple decision tree seems to be able to easily model the effect of explanations on labeling for simulated scenarios. We chose a simple learning model to establish the viability of this method, but one could easily see that the use of more sophisticated learning methods and/or more informed features should lead to better results.
\vspace{-5pt}
\paragraph{User Studies:} Next, we wanted to establish if we can still learn such simple models when the labels are collected from naive users. 
Our goal here is not to consider scenarios with possible differences in the user's knowledge, but rather cases where, even in the presence of a set of users with similar backgrounds, their responses to explanations would be too varied to learn useful models.
To test this, we used the Warehouse domain as a test bed and collected feedback on how users would view the explicability of traces generated from this domain when presented with explanatory messages detailed in Figure \ref{fig:2}.

For the study, we recruited 45 master turkers from the Amazon Mechanical Turk. Each participant was provided with the URL to a website (\url{https://goo.gl/Hun3ce}) where they could view and label various robot behaviors. We considered a setting where the robot had a full battery, but was picking up a fragile box and thus still needs to visit the station \#1. The robot could slip on some cells marked in dark grey with probability 0.25 (slipping here meant the robot picture is tilted to give an impression that it slipped on the cell and didn't prevent the robot from moving to the next cell). To make sure that all the users had similar mental models at the start, they were provided with the following facts,
(a) That robot couldn't pass through racks,
(b) Whenever the robot runs low on battery it needs to get to Station 1,
(c) Whenever the robot has a green battery sign next to the robot, that means their battery is full and
(d) The robot needs to take the shortest route to the goal.
Also, they were presented with an example trace in this instructions section and were made to take a small pre-test that allowed them to revise the above facts in various scenarios.
After the pre-test, they were shown eight traces from the robot policy sampled according to their probabilities. After the first one, the user was given an explanation message taken from the seven possible messages (the order of the messages was always randomized).

From the data collected from 45 turkers, we removed data from seven users, based on the fact they didn't find any of the transition in the first trace (i.e the case where no explanation was provided) inexplicable. We imagine this number would go down when we move to expert users or users who are invested in the success of the robot.
The data generated for the remaining 38 users were then used to train a decision tree, where the average 10-fold cross validation score for the model was at 0.935. Furthermore, we could see that the model was able to correctly predict the usefulness of intuitive minimal explanations for the given scenario. For example, it predicted that while the robots decision to visit station \#1 would be considered inexeplicable by the user in the absence of any explanation, the user would mark it as explicable when they are explained about the box being fragile and that fragile boxes need to be inspected at station \#1. In fact the model predicted that only the message that ``fragile boxes need to be inspected at station \#1" is enough to convince the user about the need for that action (i.e the user could deduce that the box must have been fragile). This shows that such learned models may help us generate cheaper explanations (the above set of explanations is smaller than the corresponding mce for the domain), by taking into account the users ability to correctly predict missing information in simple cases. Another point that surprised us was that the model predicted all slipping events as explainable even in the absence of all explanations. The cases where the user saw a slip before being told about the possibility of slipping was rare (since there are two explanatory messages related to slipping and the probability of slipping was 0.25) and furthermore when we went over the data, we found that in most such cases, the users did mark it as explainable. This may be because the effect of slipping may not have been that detrimental to the overall plan (it doesn't take you off the current path). It would be interesting to see if this result would be the same in cases where slipping was a more likely event and if it had a more apparent effect on the robot's plan.
\vspace{-2pt}
\section{Related Work}
To the best of our knowledge, this work represents the first attempt at learning standins for user mental models that allow an agent to predict the potential impact of providing explanations as model reconciliation to observers. With that said, there have been works that have looked at the problem of generating explanations in the presence of model uncertainty for human models. In particular, works like  \cite{sreedharan2018handling,sreedharan2018hierarchical} looked at cases where the agent has access to a set of potential human models. One drawback of considering a set of possible models is either they would need to have explicit sensing to identify the user model (which could mean asking a large number of questions to the user) or providing a large amount of information to cover the space of all possible models. In our work, the problem of identifying the specifics of user model is resolved through an offline training process.

Another work quite related to the discussion covered in this paper is \cite{reddy2018you}, wherein the authors tried to identify cases where they can learn a potential model for the human's expectation of the task transition dynamics when they do not align with the real world dynamics. 
Unlike their work, we do not assume that the user can provide traces for the given task, rather they may be able to provide some high-level feedback on the action (i.e. they may not be able to do or even know the right action but may be able to point out actions or transitions that surprise them). Moreover, their work requires that the user and the robot have the same reward function, which is again an assumption we do not make.
Even if we had followed their technique to learn a potential model for the human's transition function of the task, there is no guarantee that the learned representation align with the ones maintained by the human and the effects of explanatory actions may not be predictable.

\vspace{-5pt}
\section{Discussions and Conclusion}
\label{conclusions}
This paper proposes a possible way in which model reconciliation explanation could be applied to cases where the user model is unknown.
The method described here is a rather simple and general method to identify information that could potentially affect the user's mental model and produce effects that align with the agent's requirements.
There is no requirement here that the messages have to align with actual facts about the world. This again points to the rather troubling similarities between the mechanisms needed to generate useful explanations and lies \cite{Chakraborti2018WhenCA}.

Two important assumptions we made throughout the works is that the user only considers the current state (as defined by the robot) to make their decisions and we have access to a model that was learned from interactions to previous users who had similar knowledge level to the current user.
Relaxing the first assumption would require us to go beyond learning models that map transitions to labels.
We have to consider sequential models of the type considered in \cite{exp-yu} to capture the human's expectations.
As for the second, instead of assuming that all users are of the same type, a more reasonable assumption may be that the users could be clustered into N groups and we could learn a different labeling model for each user type.
Now we still have a challenge of identifying the user type of a new user and one way to overcome this would be by adopting a decision-theoretic approach to this problem and modeling it as a POMDP (where the user labels become observations and the previously learned user models turn into the observation models).

The work discussed in this paper only covers explanations that allow the user and the system to reconcile any model difference. This only covers a part of the entire explanatory dialogue. Even if there is no difference in models, the user may still have questions about parts of the policy or may raise alternative policies they think should be followed. This may arise from a difference in inferential abilities and this may require providing information that is already part of their deductive closure eg: help them understand the long term consequences of taking some actions. Once you have access to a set of such messages once could use a method similar to the one described in the paper to find the set of helpful ones. Unlike model reconciliation setting where the messages stand for information about the model, it is not quite clear how one could automatically generate such messages.

\section*{Acknowledgments} This research is supported in part by the ONR grants N00014-16-1-2892, N00014-18-1-2442, N00014-18-1-2840, the AFOSR grant FA9550-18-1-0067, and the NASA grant NNX17AD06G..

\bibliographystyle{named}
\bibliography{ijcai_subm}
\end{document}